\documentclass[letterpaper, 10 pt, conference]{ieeeconf}  

\IEEEoverridecommandlockouts                              

\usepackage{graphics} 
\usepackage{epsfig} 
\usepackage{mathptmx} 
\usepackage{times} 
\usepackage{amsmath} 
\usepackage{amssymb}  
\usepackage{siunitx} 
\usepackage{textcomp}
\usepackage{gensymb} 
\usepackage[colorlinks=true]{hyperref}
\usepackage{booktabs}
\usepackage{multirow}
\usepackage{xcolor}
\usepackage{booktabs}
\let\labelindent\relax
\usepackage{enumitem}
\usepackage{xspace}
\usepackage{booktabs}
\usepackage{soul}
\usepackage[noadjust]{cite}
\usepackage{makecell}
\usepackage{multicol}
\usepackage{rotating}
\newcommand{\myhigh}{\ul}

\DeclareMathOperator*{\argmax}{arg\,max}

\newcommand{\mypara}[1]{\par\vspace*{1.5mm}\noindent\textbf{{#1}}}
\newcommand{\etal}{\textit{et al}. }

\newcommand{\eg}{\textit{e}.\textit{g}. }

\definecolor{MyDarkBlue}{rgb}{0,0.08,1}
\definecolor{MyDarkGreen}{rgb}{0.02,0.6,0.02}
\definecolor{MyDarkRed}{rgb}{0.8,0.02,0.02}
\definecolor{MyDarkOrange}{rgb}{0.40,0.2,0.02}
\definecolor{MyPurple}{RGB}{111,0,255}
\definecolor{MyRed}{rgb}{1.0,0.0,0.0}
\definecolor{MyGold}{rgb}{0.75,0.6,0.12}
\definecolor{MyDarkgray}{rgb}{0.66, 0.66, 0.66}

\newcommand{\OURS}{AdaGrasp\xspace}
\newcommand{\OURSwoGripperSelect}{AdaGrasp-fixGripper\xspace}
\newcommand{\OURSwoConfigSelect}{AdaGrasp-fixConfig\xspace}
\newcommand{\OURSInitOnly}{AdaGrasp-initOnly\xspace}
\newcommand{\AveragePolicy}{SceneOnly\xspace}

\title{\LARGE \bf
\OURS: Learning an Adaptive Gripper-Aware Grasping Policy}

\author{Zhenjia Xu \quad Beichun Qi \quad Shubham Agrawal \quad Shuran Song  
\thanks{The authors would like to thank Lin Shao and Unigrasp authors for sharing code and models for comparison, Iretiayo A. Akinola for his help in setting up BarretHand Gripper and Google for the UR5 robot hardware.  This work was supported in part by the Amazon Research Award and the National Science Foundation under CMMI-2037101}
\\
Columbia University 
\\
\url{https://adagrasp.cs.columbia.edu}} 

\begin{document}

\maketitle
\thispagestyle{empty}
\pagestyle{empty}


\begin{abstract}
This paper aims to improve robots' versatility and adaptability by allowing them to use a large variety of end-effector tools and quickly adapt to new tools. We propose AdaGrasp, a method to learn a single grasping policy that generalizes to novel grippers. By training on a large collection of grippers, our algorithm is able to acquire generalizable knowledge of how different grippers should be used in various tasks. Given a visual observation of the scene and the gripper, AdaGrasp infers the possible grasp poses and their grasp scores by computing the cross convolution between the shape encodings of the gripper and scene. Intuitively, this cross convolution operation can be considered as an efficient way of exhaustively matching the scene geometry with gripper geometry under different grasp poses (i.e., translations and orientations), where a good ``match" of  3D geometry will lead to a successful grasp. We validate our methods in both simulation and real-world environments.  Our experiment shows that AdaGrasp significantly outperforms the existing multi-gripper grasping policy method, especially when handling cluttered environments and partial observations. Code and Data are available at \url{https://adagrasp.cs.columbia.edu}.


\end{abstract}
\section{Introduction}
In many real-world systems, a robot's end-effector is designed with a specific application in mind, where its specific geometry and kinematic structure often lead to distinct strengths and weaknesses.  
However, the vast majority of robotic research has been limited to single end-effector setups where the learned policy cannot generalize to new gripper hardware without extensive retraining. 
On the other hand, we humans can easily use various tools to accomplish different tasks and quickly adapt to unseen tools. Can we allow our robot system to do the same? 
This capability would benefit a robot manipulation system in the following ways:
\begin{itemize}[leftmargin=*]
    \item \myhigh{Versatility via diversity.} Since different gripper designs often provide complementary strengths and weaknesses, and by learning to adequately use \textit{a diverse set of} grippers, the system can effectively improve its versatility on handling a larger variety of objects and tasks.

    \item \myhigh{Adaptability via generalization.} Since the learned grasping policy can generalize across different gripper hardware, it can also quickly adapt to \textit{new} grippers by directly analyzing its geometry and structure.  It is different from the existing multi-gripper systems \cite{Zeng2018,mahler2019learning} that need to collect new training data for any new gripper hardware. 
\end{itemize}

To achieve this goal,  we propose \textbf{\OURS}, a learning-based algorithm that learns a unified policy for different grippers and can generalize to novel gripper designs.
At its core, \OURS uses cross convolution (CrossConv)\cite{visualdynamics16} operation between the shape encoding of the robot gripper and the scene to infer the grasp score for all possible grasp poses. Intuitively, this operation can be considered as an efficient way of exhaustively matching the scene and the gripper geometry under different grasp poses, where a good ``match'' of their 3D geometry will lead to a successful grasp.

\begin{figure}[t]
    \centering\vspace{-1mm}
    \includegraphics[width= 0.97\linewidth]{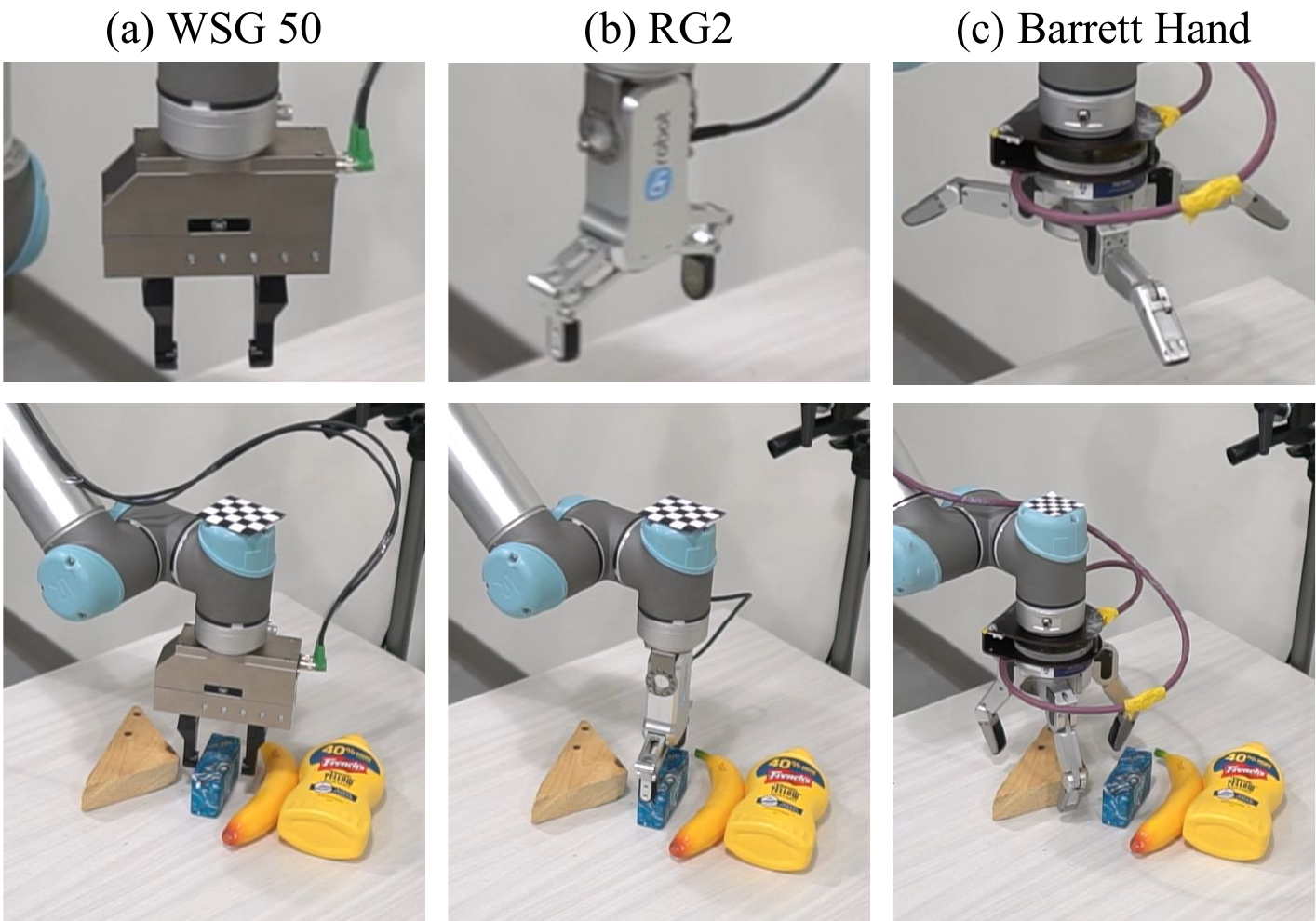} \vspace{-3mm}
    \caption{ \textbf{Gripper-Aware Grasping Policy.}
    The goal of \OURS is to produce grasping strategies that are conditioned on input gripper description (a,b,c). 
    For example, since the RG2 gripper has a wider fixed opening than WSG 50 (which can control its opening width), it chooses a different grasp pose to avoid double-picking or collision. Barrett Hand grasps the big triangle shape, which can be challenging for other two-finger grippers.}
    \label{fig:example}
    \vspace{-6mm}
\end{figure}

\begin{figure*}[t]
    \centering
    \includegraphics[width=0.96\linewidth]{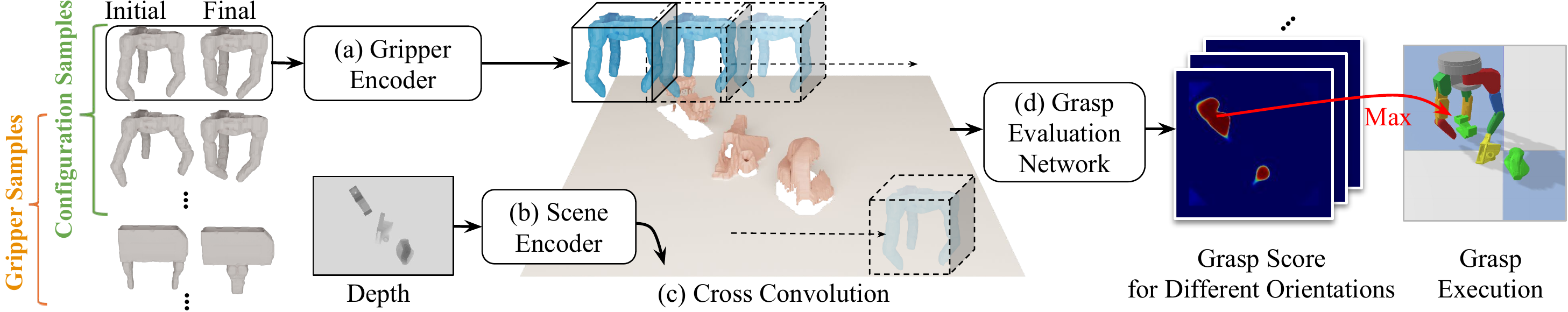}  \vspace{-4mm}
    \caption{\textbf{Approach Overview.} At its core, \OURS infers grasp scores for all candidate grasp poses by computing the cross convolution between the gripper encoding produced by the gripper encoder (a) and the scene encoding produced by the scene encoder (b). This cross convolution operation (c) matches the scene and gripper encoding under all grasp poses by translating and rotating the gripper kernel,  where a good ``match'' of their encoding results in a high grasp score. Different initial opening configuration of a gripper are treated as different grippers and fed to the network in parallel.  The action associated with the highest grasp score is executed. The action for a grasp attempt includes selecting the suitable gripper, deciding its initial joint configuration, and choosing a proper grasp pose. }
    \label{fig:network}
    \vspace{-6mm}
\end{figure*}

The 3D geometry of a robot gripper and its kinematic structure often inform how it should be used for a given task \cite{ha2020fit2form}. By learning to use a large collection of different grippers, the algorithm should be able to acquire a generalizable knowledge of how different grippers should be used in various tasks. 
For example, a gripper's opening width determines what object shape can fit into the gripper, and the thickness of each finger determines what narrow space the finger can get into without collision.
Fig.\ref{fig:example} illustrates different grasp poses that are suitable for different grippers.

The primary contribution of this paper is \OURS, a learning-based grasping algorithm that leverages generalized shape matching via cross convolution to produce a grasping policy that works across different gripper hardwares.  We validate our methods in both simulation and real-world environments. Our experiments show that \OURS outperforms the state-of-the-art method for multi-gripper grasping, especially in a cluttered environment and with partial observation.





\section{Related Work}
\mypara{Learning-based single-gripper systems.}
%
Recent data-driven methods have made great progress on learning object-agnostic grasping policies that detect grasps by exploiting visual features, without explicitly using object-specific prior knowledge \cite{song2020grasping, redmon2015real,pinto2016supersizing,gualtieri2016high, lu2018planning,mousavian20196,gualtieri2018learning, morrison2018closing, choi2018softrobothand, yan2018geometryaware, liang2019pointnetgpd, murali2020clutter}.
These algorithms demonstrate the ability to generalize to new objects and scene configurations.
However, they are often designed and trained with a fixed hardware setup. Hence, they cannot adapt to any changes in the gripper hardware without extensive retraining.   

\mypara{Learning-based multi-gripper systems.} To take advantage of complementary skills between different grippers, more recent works have started to use multiple end-effectors for grasping. For example, both Zeng \etal \cite{Zeng2018,zeng2016multi} and Mahler \etal \cite{mahler2018dex} used a setup with one suction cup gripper and one parallel jaw gripper. 
However, in both the systems, the algorithm learns a separate policy for each gripper, i.e., their policies cannot generalize to new grippers.  As a result, these algorithms are often limited to a small number of grippers. 

\mypara{Contact-based grasping policy.}
Many analytical grasping models have been proposed to evaluate grasp quality through contact-point reasoning and force-closure analysis \cite{varley2015generating,veres2017modeling,le2010learning,varley2017shape,mahler2016dex}.  
The work most related to us is UniGrasp \cite{shao2020unigrasp}, where the algorithm takes in the gripper point cloud and a single object point cloud, samples N points from the object point cloud as contact points for N fingers, and uses inverse kinematics to get gripper joint configuration.

While a contact-based policy generalizes to new grippers, it also brings in limitations. 
First, since measuring precise contact points in real-world is challenging, the algorithm can only be trained with simulation. Moreover, it is trained using static force closure analysis, which does not consider the object dynamics during grasping.
Second, to reason about force closure, the algorithm assumes a complete object representation as input which relies on a perception algorithm to perfectly detect the target object and provide full 3D geometry. Since the algorithm only samples contact points on the object surface, a partial observation of the object will lead to unstable contact point selection and inaccurate force closure evaluation, as we showed in our experiments. In contrast, our method's action space won't be limited by partial observation.
Furthermore, it does not consider the gripper geometry beyond contact points, which increases the likelihood of collision in cluttered environments.  
In contrast, our algorithm does not require any explicit contact point supervision or complete object representation. Therefore, it can better handle cluttered environments and partial observation.

\section{Approach}
The goal of our algorithm is to learn a policy that can produce the optimal grasping strategy for a novel gripper by estimating the probability of grasp success (i.e., grasp score) for all candidate gripper configurations and grasp poses. 
Concretely, taking a visual observation of the scene (RGB-D images) and the gripper design (defined as URDF files) as input, the algorithm infers the possible grasp poses along with their grasp scores that would allow the gripper to successfully grasp a target object.

The core of our approach is a Grasp Evaluation Network $f_{grasp}(s, g) \rightarrow a $ that infers the grasp score for all candidate grasp poses $a$ by computing the cross convolution between the gripper encoding $g$ and scene encoding $s$. The grasp pose is parameterized by rotation about the z-axis and 2D translation. 
This cross convolution operation can be considered as an efficient way of exhaustively matching the scene geometry with gripper geometry in all possible grasp poses by translating and rotating the gripper kernel. The matching score is finally represented as a dense grasp score map, where a higher value indicates a higher chance of a successful grasp. We train the algorithm with a collection of grippers and environment setups and test it with unseen grippers and objects. Fig. \ref{fig:network} shows the network overview, and the following sections provide details of our approach.

\subsection{Gripper and Scene Representation}


\mypara{Gripper encoding}.
The gripper geometry is captured by 10 depth images and encoded as a 3D TSDF volume \cite{newcombe2011kinectfusion}.  The volume dimension is $64\times 64\times 32$ (voxel) with voxel size $v_g=0.004$ (m).
We compute TSDF volume for the gripper at its initial open state and final closed state and stack them as input $I_g\in R^{2\times 64\times 64\times 32}$. The gripper encoder network (Fig \ref{fig:network} a) starts with two 3D convolution layers with kernel size $3\times3\times3$, resulting in a feature $\in R^{64\times32\times32\times16}$.
Then we use one 3D convolution with kernel size $1\times1\times16$ reducing the z dimension to 1. 
Finally, we use 5 2D convolution layers to produce the gripper features $\psi(g) \in R^{16\times 32\times 32}$. 

\mypara{Scene encoding} 
The input scene is captured with a top-down depth image and encoded as a 3D TSDF volume. 
The workspace dimension is $192\times 192\times 64$ (voxel) with a voxel size $v_s=0.002$ (m).
In multi-object obstacle cases, the obstacle mask is provided as an additional channel. This channel will be 0 for other cases. 
The scene volume $I_s\in R^{2\times 192\times 192\times 64}$ is then fed into the scene encoder network (Fig \ref{fig:network} b). Similar to the gripper encoder network, it consists of three 3D convolution layers with downsample scale=4, one layer for z-axis reduction, and five 2D convolution layers. The output is the scene features $\phi(s) \in R^{16\times 48\times 48}$.


\begin{figure}[t]
    \includegraphics[width=0.99\linewidth]{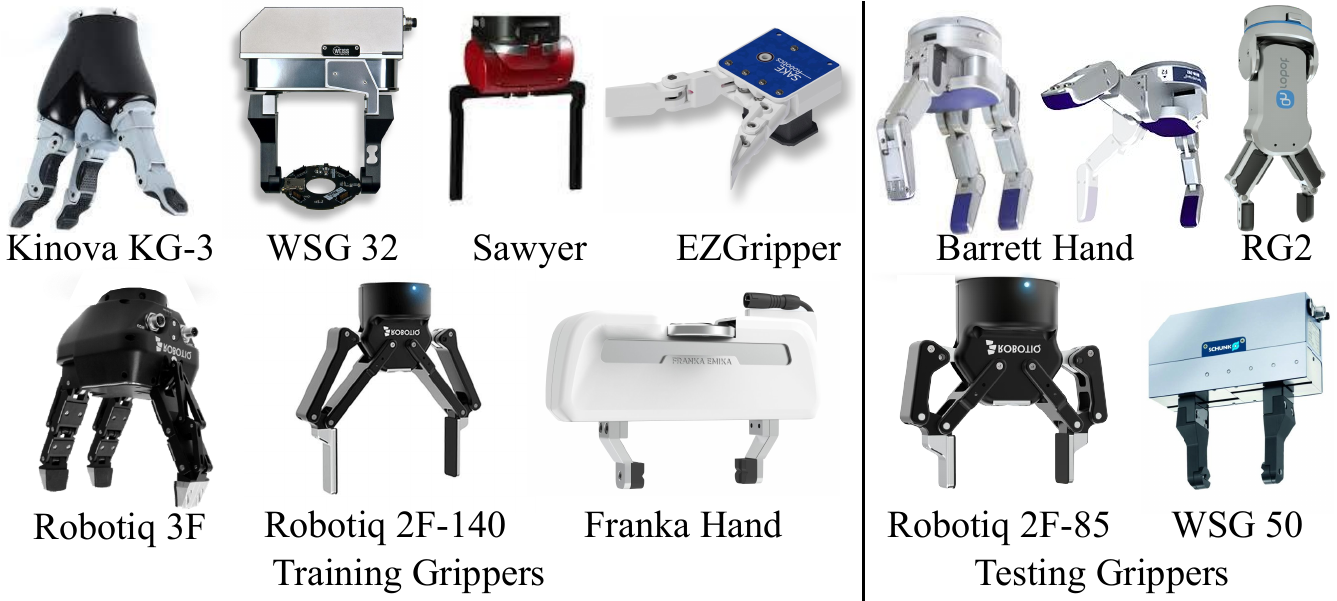} \vspace{-2mm}
    \caption{Training and testing grippers used in our experiments.}
    \label{fig:grippers}
    \vspace{-6mm}
\end{figure}

\subsection{Grasp Evaluation via Shape Matching} 
After the encoding network, the scene and gripper geometry are mapped into a query $\phi(s)$  and key  $\psi(g)$ features. We carefully set the number of downsampling size in scene encoder and gripper encoder so that both features share a similar physical receptive field. 
As a result, the spatial alignment is maintained, and shape matching in feature space (via CrossConv) is meaningful. 
The algorithm then computes the cross convolution between the $\psi(g)$ and $\phi(s)$ by treating $\psi(g)$ as the convolution kernel (Fig. \ref{fig:network} c). The output shares the same size as the scene feature $\phi(s)$. We repeat this step for $r=16$ times \cite{Zeng2018a}, each time rotate the scene TSDF volume by $\theta = 2\pi/r$ about z-axis.  
Finally, the output of cross convolution is fed into a grasp evaluation network (Fig \ref{fig:network}-d) that estimates dense grasp scores for all possible actions $ Q \in R^{X_s\times Y_s \times r} $, where each grasp score $Q(i,j,k)$ in the $Q$ value map corresponds to one grasp pose. 

The grasp pose is parameterized by its position $(x,y,z)$ and orientation $\theta = k\pi/r$ about z-axis, where $ x = x_{min}+v_si$, $y = y_{min}+v_sj$, $z=\mathcal{H}(O(i,j))-0.05$, $[x_{min}, y_{min}, z_{min}, x_{max}, y_{max}, z_{max}]$ is the workspace bound, $\mathcal{H}(O(i,j))$ is the height of z-dimention in the scene volume $O$ at location $(i,j)$. During grasp execution, the gripper starts at location $(x,y,z_{max})$, moves downward along z-axis until having contact with an object or reaching the target position $(x, y, z)$, and then close its finger. The gripper will then move upwards and this execution is considered successful if and only if exactly one target object is lifted $>0.2$m. Grasping an obstacle or more than one objects is classified as a failure.

\mypara{Network training.} 
The whole network is trained end-to-end with self-supervised grasping trials, similar to prior work \cite{song2020grasping,Zeng2018a}. Based on the object height after grasping, each grasp trial is labeled with its grasp outcome (1 = success, 0 = failure). The network is trained to predict the grasp outcome for all possible actions, and it is supervised by the grasping outcome of the executed action (one action out of $X_s\times Y_s\times r$ actions) using softmax loss. 

During training, the network chooses its action using  $\epsilon-$greedy. We use the normalized predicted grasp scores as the probability of choosing each pose.  
At training epoch $e$, $\epsilon$ decreases linearly from $\epsilon_{max}$ to $\epsilon_{min}$. After $n$ epochs, $\epsilon=\epsilon_{min}$. We set $n=2000, \epsilon_{min}=0.2, \epsilon_{max}=0.8$. 
All the grasp trails are stored in a FIFO replay buffer (size=12000). At each training step, we sample a batch of examples from the replay buffer with a 1:1 positive to negative ratio.
We also used data augmentation to overcome overfitting. The scene inputs obtained from the replay buffer have a probability of 0.7 to be randomly shifted and rotated. We applied the same transformation to the corresponding grasp pose.
The final model is trained for 5000 epochs, 8 sequences of data collection, and 32 iterations of training per epoch with Adam optimizer and learning rate 0.0005.


\begin{table*}[t]\vspace{-1mm}
    \centering
    \setlength\tabcolsep{5.4pt}
    \caption{{Grasp Success Rate.} \label{tab:result} } \vspace{-3mm}
    \begin{tabular}{l|cccc|cccc|cccc}
        \toprule
        Algorithm &  \multicolumn{4}{c|}{Single object} & \multicolumn{4}{c|}{Multi-object} & \multicolumn{4}{c}{Multi-object w. obstacles} \\
        &$O_{tr}\textrm{-}G_{tr}$
        &$O_{tr}\textrm{-}G_{te}$
        &$O_{te}\textrm{-}G_{tr}$
        &$O_{te}\textrm{-}G_{te}$
        &$O_{tr}\textrm{-}G_{tr}$
        &$O_{tr}\textrm{-}G_{te}$
        &$O_{te}\textrm{-}G_{tr}$ 
        &$O_{te}\textrm{-}G_{te}$
        &$O_{tr}\textrm{-}G_{tr}$
        &$O_{tr}\textrm{-}G_{te}$
        &$O_{te}\textrm{-}G_{tr}$ &$O_{te}\textrm{-}G_{te}$\\

        \midrule
        \AveragePolicy  & 0.613 & 0.730 & 0.596 & 0.686 & 0.493 & 0.528 & 0.497 & 0.531 & 0.347 & 0.368 & 0.271 & 0.303 \\
        SingleGripper \cite{Zeng2018a} & - & 0.930 & - & 0.930 & - & 0.788 & - & 0.886 & - & - & - & - \\

        UniGrasp \cite{shao2020unigrasp}  & - & - & - & 0.812 & - & - & - & 0.228 &  - & - & - & - \\

        
        \midrule
        
        \OURSInitOnly & 0.721 & 0.792 & 0.719 & 0.753 & 0.637 & 0.674 & 0.638 & 0.626 & 0.494 & 0.353 & 0.513 & 0.343 \\
        
        \OURSwoConfigSelect & 0.766 & 0.855 & 0.770 & 0.854 & 0.706 & 0.751 & 0.685 & 0.764 & 0.612 & 0.523 & 0.603 & 0.569 \\
        
        \OURSwoGripperSelect & 0.923 & 0.905 & 0.959 & 0.938 & 0.849 & 0.842 & 0.875 & 0.854 & 0.775 & 0.658 & 0.813 & 0.703 \\
        
        \OURS & \textbf{0.960}& \textbf{1.000} & \textbf{0.970} & \textbf{0.990} & \textbf{0.912} & \textbf{0.908} & \textbf{0.896} & \textbf{0.936} & \textbf{0.853} & \textbf{0.747} & \textbf{0.887} & \textbf{0.793} \\

     \bottomrule
    \end{tabular}
    \begin{center}
    \vspace{-1mm}
        Test case is labeled by $O_\textrm{object\ type}$-$G_\textrm{gripper\ type}$ (tr: train, te: test). Note: UniGrasp is tested with 4-camera input, all others are tested with 1-camera input.
    \end{center}
    
    \vspace{-8mm}
\end{table*}

\subsection{Improving Grasp Quality via Gripper Selection}
To execute the grasp, the algorithm selects the predicted best action from the grasp evaluation network  $ a = \argmax_a Q $. However, depending on the input gripper, sometimes even the best action might still not be good enough to achieve a successful grasp (e.g., the input gripper or its initial configuration is too small to enclose the object inside). 
In such cases, the algorithm will compare and select between different input grippers to improve its grasp quality.

To do so, the network predicts a grasp score for a list of $N$ candidate grippers, then selects the one that produces the highest grasp score. Note that the list of candidate grippers can include \textit{completely different grippers} or \textit{the same gripper with different initial joint configurations}. 
Since the grasp evaluation network is trained for many grippers, the estimated grasp score for different grippers is naturally comparable, where a higher score indicates a better gripper for the task. 
During testing, we allow the algorithm to choose the best configuration for a given gripper (\OURSwoGripperSelect in Tab. \ref{tab:result}) or choose both the best gripper and its best configuration at the same time (\OURS  in Tab. \ref{tab:result}).

\mypara{Configuration Sampling.} To sample possible initial configuration for a given gripper, we linearly map the gripper's joint configuration into a scalar value in the range [0,1], where 0 represents the fully closed state, and 1 represents the fully open state. 
Note that the algorithm only needs to choose grippers' initial configuration, since the final configuration is determined -- the gripper will always try to close its fingers all the way to its fully closed state. 

During training, each gripper has 4 initial configuration options randomly sampled between 0.4 and 1.0. 
Since two fingers of Barrett Hand have flexible palm joints, we define the following 3 presets:
(1) palm joint = 0, two flexible fingers are parallel and next to each other.
(2) palm joint = $0.1\pi$, the angle between two flexible fingers is $0.2\pi$.
(3) palm joint = $0.5\pi$ and remove the finger with a fixed palm joint. This configuration mimics a broken Barrett Hand with  two remaining fingers (Barrett Hand-B).


\section{Experiments}
We run the following experiments to verify that the proposed \OURS algorithm is able to (1) learn different grasping strategies for different grippers, (2) generalize to new grippers, (3) select a suitable gripper and gripper configuration for a given task. We have also provided real-world experiments to validate our approach. 

\mypara{Scene setup:}
We use Pybullet \cite{coumans2016pybullet} as our simulation environment. The target objects and obstacles are randomly dropped within a rectangular workspace. 
All objects used in simulation are from Dexnet 2.0 \cite{mahler2017dex} object dataset. 
The training dataset has 801 objects: 400 from the 3DNet subset and 401 from the Kit subset. The test dataset has 57 objects: 13 from Adversarial subset and the remaining object from the Kit category that are not used in training. 

For our method, we use a single top-down RGB-D camera to capture the scene. For UniGrasp, we use 3 additional cameras to provide a complete 3D point cloud input since it is sensitive to partial observation. Tab. \ref{tab:visibility} studies both algorithm's performance with respect to scene visibility. 
We tested the following scenarios:
\begin{itemize}[leftmargin=*]
\item Single object. One random object is dropped into the scene with random position and orientation.
\item Multiple objects. There are 5 objects in the scene, and the gripper is expected to grasp one object at a time until the scene is empty or a maximum attempt of 7 is reached.  
\item Multiple objects with obstacles. There are 3 targets and 3 obstacles. We provide the obstacle mask. The algorithm needs to grasp the target object while avoiding obstacles. 
\end{itemize}

\mypara{Gripper:}
We have 7 training grippers and 4 testing grippers as shown in Fig. \ref{fig:grippers}. One of the testing grippers is Barrett Hand with one finger missing, which is equivalent to a 2 finger gripper. During training, grippers are globally scaled by a random factor of $t\in (0.8,1.2)$ to increase the training gripper diversity. During testing, gripper scale is fixed at 1.

\mypara{Metric:} The algorithm performance is measured by grasp success rate = $\frac{\mathrm{\# successful\_grasps}}{\mathrm{\# total\_grasp\_attempts}}$. The grasp success for each attempt is measured by whether the gripper grasps strictly one target. For example, in the multi-object setup, grasping two objects simultaneously is considered a failure (double-picking). The objects can be grasped in any order.  

We evaluate the algorithms on all grippers separately and use the average performance, except in our final policy, the algorithm has the freedom to select from a set of grippers. For each type of scene, the test scene generation is consistent across all algorithms and grippers.

\mypara{Algorithm comparisons:} 
\begin{itemize}[leftmargin=*]
\item UniGrasp \cite{shao2020unigrasp}: it takes in the gripper point cloud and object point cloud (background removed), samples N (2 or 3) points from the object as contact points for N fingers, respectively, and use inverse kinematics to compute gripper joint configurations for grasp execution. We directly test the pre-trained model provided by the authors.
\item \AveragePolicy: a single policy trained using all training grippers (uniformly sampled during training). The policy can only access the scene observation without gripper information; hence, it predicts uniformly across all grippers.
\item SingleGripper \cite{Zeng2018a}: a learning based grasping method from Zeng et al. using only Robotiq 2F-85.
\item \OURSInitOnly: the gripper input is the initial gripper state. The policy selects the best grasp pose (position and orientation) for a given gripper.
\item \OURSwoConfigSelect: same as \OURSInitOnly, but gripper input has both its initial and final state.
\item \OURSwoGripperSelect: the algorithm linearly samples the gripper configurations and infer grasp score for each configuration. Then, the algorithm selects the gripper configuration with the highest grasp score to execute. 
\item  \OURS: On top of the gripper configuration and grasp pose, this algorithm also selects the best gripper with the highest grasp score to use. This is our final policy.
\end{itemize}
In testing, \AveragePolicy, \OURSInitOnly, and \OURSwoConfigSelect uses a random initial configuration sampled from [0.5,  0.625,  0.75,  0.875,  1.0]; \OURSwoGripperSelect and \OURS will select the configuration from the same list.

\subsection{Experimental Results} 
\begin{figure}[t]
    \centering \vspace{-1.5mm}
    \includegraphics[width=0.95\linewidth]{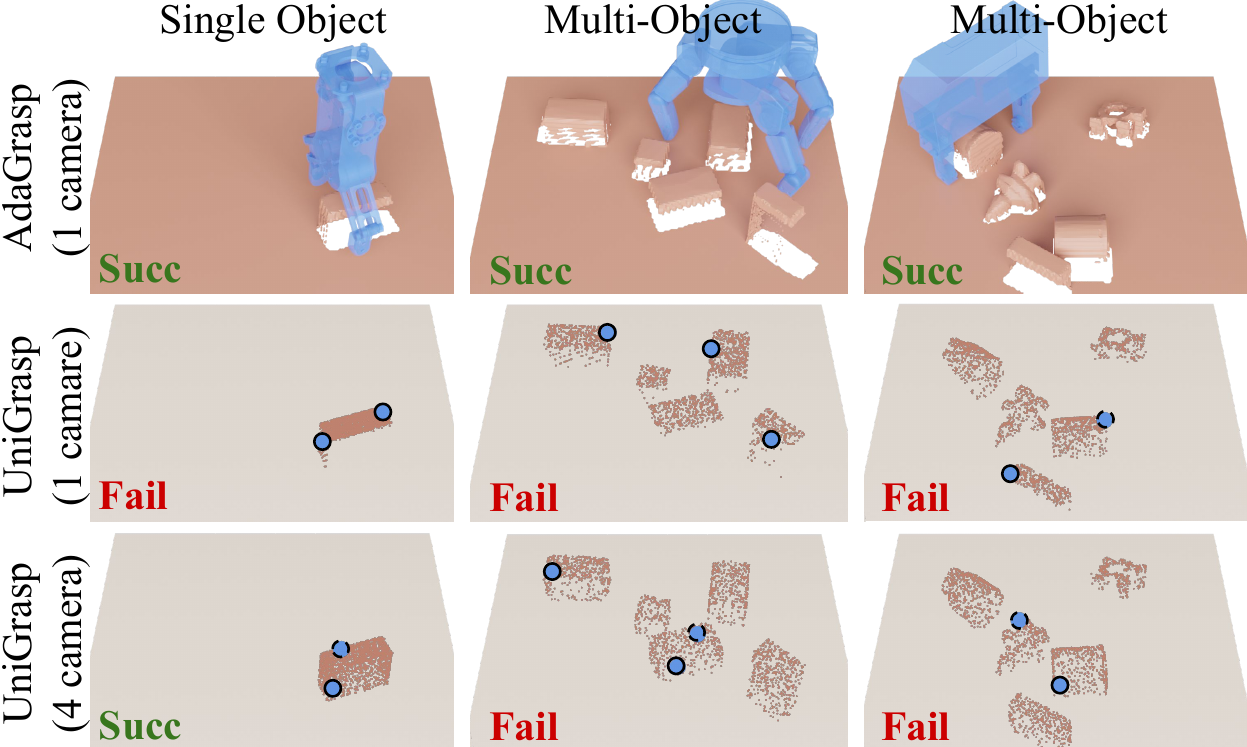} \vspace{-3mm}
    \caption{ \textbf{Comparisons. }
    UniGrasp often fails on incomplete input point clouds since it samples contact points directly from the pointcloud (2nd row with 1 camera). It also struggles with cluttered scenes, frequently sampling contact points on multiple objects or failing to account for collision. \OURS is able to handle both partial observability and scene clutter. }
    \label{fig:comparison} \vspace{-3mm}
\end{figure}

\mypara{Comparison to prior work.}
We compare our approach with state-of-the-art multi-gripper system UniGrasp \cite{shao2020unigrasp}. The number of cameras during AdaGrasp's training is randomly chose in \{1,2,3,4\}. Both algorithms are evaluated on test objects and test grippers under a fixed-gripper and fixed-camera setting (i.e., the algorithm can choose the input gripper's initial configuration but cannot switch gripper).
In the single object case,  \OURSwoGripperSelect  achieves better performance (+10\%) comparing to UniGrasp.
The advantage is much more salient in multi-object case, where \OURSwoGripperSelect is able to outperform UniGrasp by around 60\%. This result highlights \OURS's ability in handling cluttered environments. Fig. \ref{fig:comparison} shows qualitative comparisons, where UniGrasp samples contact points on multiple objects or misses potential collisions.

Another advantage of \OURS is its ability in handling partial observations. UniGrasp is very sensitive to the quality and visibility of scene observation since it directly samples contact points from the input point cloud, which is limited to the observed surface (Fig. \ref{fig:comparison}-a). In contrast,  \OURS is able to reason about the object grasp point beyond the visible surfaces using 3D TSDF representation. Results in Tab. \ref{tab:visibility} demonstrate that when the scene observation is incomplete (i.e., with fewer cameras), UniGrasp's performance decreases significantly, while \OURS has consistent performance.
Inference time of \OURS is $1.05$s for each gripper with 5 initial configurations and 16 rotations. 

\begin{table}[t]
\caption{Grasp Succ Rate w.r.t Partial Observation. \label{tab:visibility}}   \vspace{-3mm}
    {\footnotesize
    \centering
    \setlength\tabcolsep{2pt}
    \begin{tabular}{c|cccc|cccc}
    \toprule
     & \multicolumn{4}{c|}{Single Object} & \multicolumn{4}{c}{Multi Object} \\
    \# Camera & 4 & 3 & 2 & 1 & 4 & 3 & 2 & 1 \\
    \midrule
    UniGrasp \cite{shao2020unigrasp} 
    & 0.812  & 0.788  & 0.768  & 0.732 
    & 0.228  & 0.258 &  0.254 & 0.175 \\
    \OURSwoGripperSelect 
    & 0.896 & 0.892 & 0.889 & 0.891 & 0.854 & 0.863 & 0.846 & 0.821 \\
    \bottomrule
    \end{tabular}}
    
\vspace{1mm}
The success rate of UniGrasp degrades as the number of cameras and scene visibility decreases, whereas \OURS performs consistently throughout. Both algorithms are tested with our test objects and test grippers under a fixed-gripper setting. 
    \vspace{-7mm}
\end{table}

\begin{figure*}[t]
    \centering  
    \includegraphics[width=0.99\linewidth]{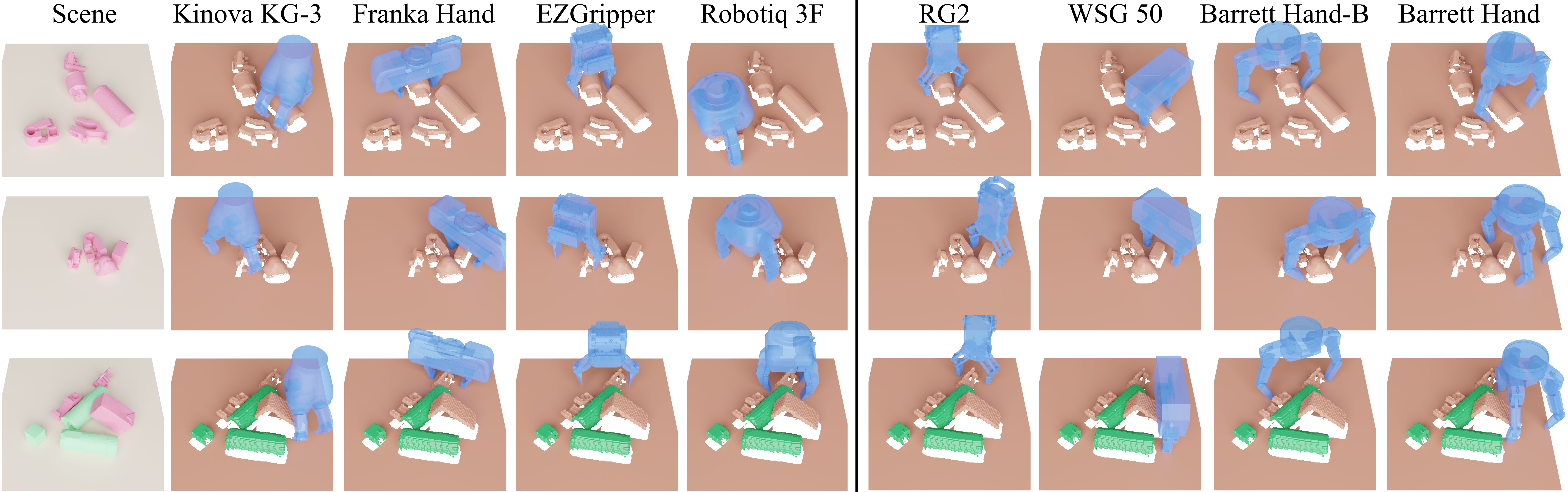}\vspace{-3mm}
    \caption{ \textbf{Gripper-Aware Grasping Policy. } Given the same input scene in each row, \OURS predicts a different grasp pose suitable for each gripper. Here are example grasps inferred by \OURS for training grippers (left) and testing grippers (right) in multi-object setups (Row 1-2), and multi-object + obstacle setups (Row 3).  Brown surface: input TSDF. Green surface: obstacles input as additional mask. More examples available on our website. \label{fig:action_result}} 
\end{figure*}

\begin{figure*}[t]
    \centering 
    \includegraphics[width=0.99\linewidth]{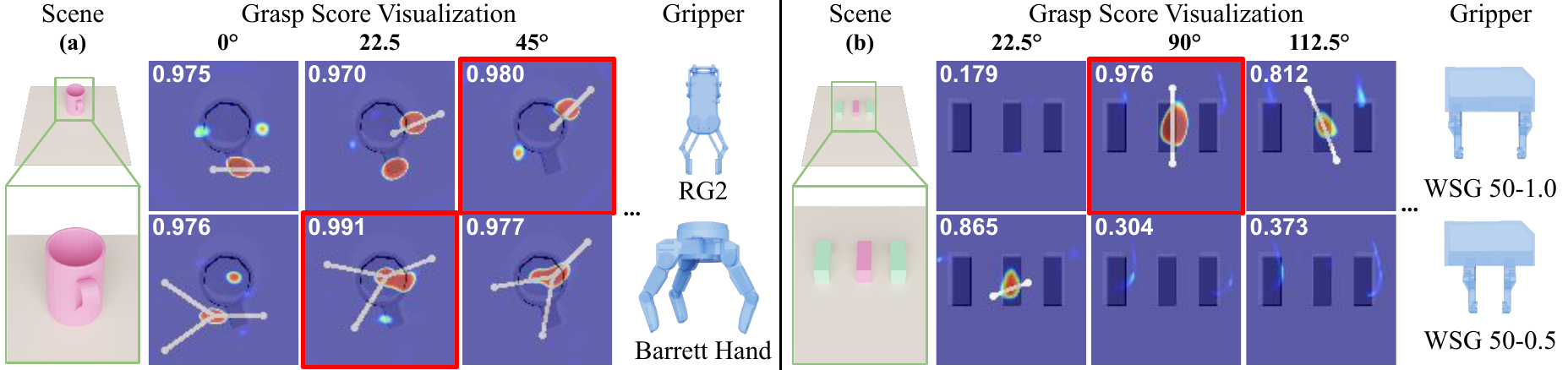} \vspace{-3mm}
    \caption{\textbf{Grasp Score Visualization.} Dense grasp score predictions are shown for 3 out of 16 different grasp orientations. The highest grasp score for each orientation is shown at the top left. For each gripper, the orientation with the highest score  is highlighted in red. 
    In scene (a), the target object is a mug. RG2 prefers to grasp the cup's edge or handle, while Barrett Hand prefers to grasp across the whole cup. 
    In scene (b), the target object is surrounded by two obstacles (green). We visualize the grasp poses for the WSG 50 gripper under different initial configurations (opening size). With a larger opening, the algorithm chooses to grasp vertically (90\degree) to avoid collisions, while with a smaller opening, it chooses to grasp horizontally  (22.5\degree) since the object's length is now larger than the gripper width. Between these two configurations, the algorithm chooses the wider opening. 
    } \vspace{-5mm}
    \label{fig:score_result}
\end{figure*}

\mypara{Can \OURS learn gripper-aware grasping policy?}
 To verify \OURS's ability to infer different grasping strategies conditioned on the input gripper, we perform the following experiments. All models in Tab. \ref{tab:result} are trained and tested under single-camera setting. First, we compare \OURSwoConfigSelect with an ``\AveragePolicy'' policy, i.e., a single policy trained with all training grippers without the gripper as input. Results in Tab. \ref{tab:result} shows that \OURSwoConfigSelect's performance is always significantly better than the ``\AveragePolicy'', which demonstrates that \OURSwoConfigSelect improves the grasp prediction by analyzing the input gripper. 
 We visualize the top grasp pose prediction for different grippers given the same scene setup (Fig. \ref{fig:action_result}  \ref{fig:score_result}). From the visualization, we can see that the algorithm is able to infer diverse grasp poses that are suitable for each input gripper and configuration. 

 \begin{figure}[t]
    \centering
    \includegraphics[width=0.98\linewidth]{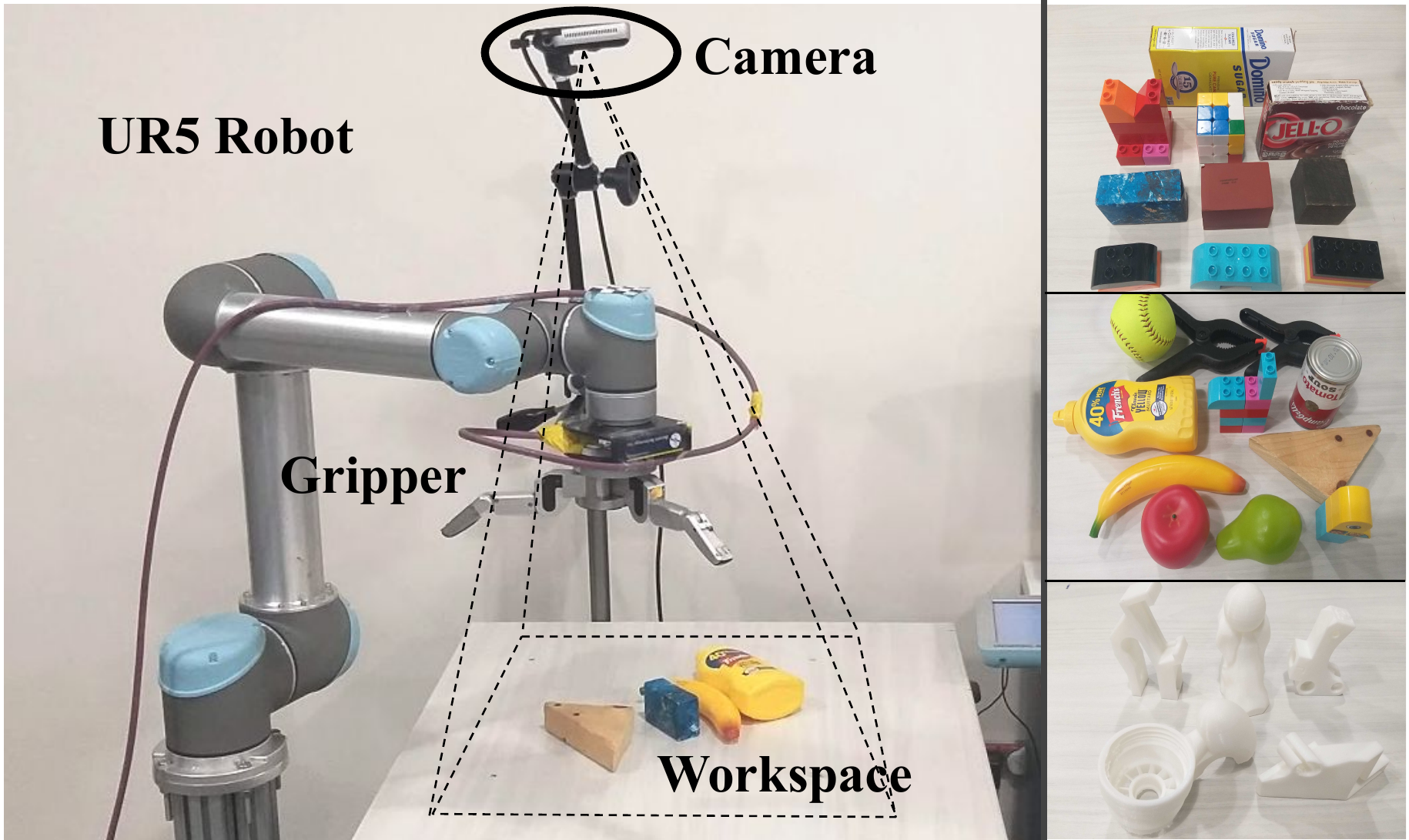} \vspace{-3mm}
\caption{\textbf{Real-world Setup.} Robot and camera setup (left) and test objects (right). Videos of experiments are available in supp. video and website. \label{fig:real-setup}}  \vspace{-2mm}
\end{figure}

\begin{table}[t]
    \centering
    \caption{Real-world Grasp Succ Rate on Unseen Grippers and Objects.} \vspace{-3mm}
    \begin{tabular}{l|cccc}
    \toprule
         & WSG 50  & RG2 & Barrett Hand & Barrett Hand-B \\
    \midrule
    Single Object & 0.92 & 0.92 & 0.88 & 0.80 \\
    Multiple Objects & 0.90  & 0.88 & 0.78 & 0.66 \\
    \bottomrule
    \end{tabular}
    \label{tab:real}  \vspace{-8mm}
\end{table}

\mypara{Can \OURS generalize to new grippers?} 
To test the algorithm's adaptability to new gripper hardware, we tested the learned policy with five unseen grippers, including three 2-finger grippers, one 3-finger grippers, and a ``damaged'' 3-finger gripper (Barrett hand with a missing finger). While test grippers are never used during training, \OURSwoGripperSelect is able to get performance comparable to that on the training grippers. In Tab. \ref{tab:result} \OURSwoGripperSelect improves the \AveragePolicy policy performance by 18\% to 54\%. 

\mypara{Can \OURS select the right configuration and gripper for a given task?} 
To check whether the predicted grasp score is informative for comparing and selecting the gripper's initial configuration, we compare the algorithm performance with and without configuration selection (\OURSwoConfigSelect v.s. \OURSwoGripperSelect). Both algorithms predicts the grasp scores for the same gripper. The difference is that \OURSwoGripperSelect selects the configuration with the highest grasp score while \OURSwoConfigSelect randomly picks one configuration. Compared to \OURSwoConfigSelect, \OURSwoGripperSelect performance is better in all cases, improving 5\% to 21\%. This result validates that the predicted grasp score is informative for selecting the best initial configuration. Fig. \ref{fig:score_result}-b shows an example of configuration selection for WSG 50. 

Similarly, we showed that the grasp score is also comparable across different grippers. As a result, the algorithm is able to further improve its grasping performance by choosing the ``right tool'' (gripper)  for a given task at hand (object to grasp).  Comparing \OURS with \OURSwoGripperSelect in Tab. \ref{tab:result}, we can see the 1\% to 9\% improvement in all scenarios. The performance of \OURS is also better than SingerGripper, which only evaluates on Robotiq 2F-85. This result indicates that if combined with an automatic tool changing hardware \cite{toolchanger}, \OURS can  improve the grasping performance by allowing the system to properly use a diverse set of grippers.

\mypara{Is gripper final state encoding helpful?}
The input gripper encoding includes both gripper's initial and final state. It allows the algorithm to reason about the gripper's dynamics during the closing action beyond its static 3D geometry. To see the effect of final state encoding, we compare the model without the final-state, which is \OURSInitOnly. In almost all test cases, \OURSwoConfigSelect has a higher success rate, and it is most salient in the multi-object with obstacles setup (up to +23\% improvement). Moreover, \OURSwoConfigSelect demonstrates better generalizability when testing on new gripper hardware. 

\mypara{Real-robot experiment}
Finally, we validate our method on a real-world robot platform with a UR5 robot and a calibrated RGB-D camera (Intel RealSense D415).  
Fig. \ref{fig:real-setup} shows the real-world setup and test objects. 
In this experiment, we directly tested \OURSwoGripperSelect policy trained in simulation on four different physical grippers -- WSG 50, RG2, Barrett Hand, and Barrett Hand-B, all of which are unseen during training. The test objects used in this experiment include 20 objects from YCB dataset \cite{calli2017yale} and five 3D printed adversarial objects from DexNet 2.0, all unseen during training.
For single object tests, we place a single object randomly. For multi-object tests, we created 8 scenes each containing 4 randomly chosen objects and made sure that the placement of objects in 8 scenes is consistent across grippers for fair comparison. For each multi-object scene, we provide 7 attempts to a gripper for grasping objects. The grasp success rates are reported in Tab. \ref{tab:real}. The average success rates for single object and multi-object are 86\% and 80.5\%, respectively, comparable with the algorithm performance in simulation. We noticed that unlike parallel jaw grippers, Barrett Hand and Barrett Hand-B have a curved grasping gait, i.e., fingers take a curved trajectory while closing in. Thus, the Barrett Hand cannot create contact at a smaller height and fails to grasp shorter objects like banana and adversarial objects. On the other hand, Barrett Hand is good at grasping bigger objects like big triangle or baseball ball, which are challenging for smaller grippers like RG2.

\section{Conclusion and Future Directions}
We introduced  \OURS, a unified policy that generalizes to novel gripper designs.  
Extensive experiments demonstrate that \OURS is  able to improve the system's versatility and adaptability, and outperforms the current state-of-the-art multi-gripper grasping method. 
However, since our algorithm focuses on the gripper geometry for mechanical gripper, it does not extend to other gripper types (e.g., suction or deformable) and variable physical parameters (\eg friction). It is also limited to top-down grasps due to the reduced action space. As future directions, it will be interesting to investigate larger range of gripper types in general dexterous manipulation.




\bibliographystyle{IEEEtran}
\bibliography{reference}

\begin{thebibliography}{10}
\providecommand{\url}[1]{#1}
\csname url@rmstyle\endcsname
\providecommand{\newblock}{\relax}
\providecommand{\bibinfo}[2]{#2}
\providecommand\BIBentrySTDinterwordspacing{\spaceskip=0pt\relax}
\providecommand\BIBentryALTinterwordstretchfactor{4}
\providecommand\BIBentryALTinterwordspacing{\spaceskip=\fontdimen2\font plus
\BIBentryALTinterwordstretchfactor\fontdimen3\font minus
  \fontdimen4\font\relax}
\providecommand\BIBforeignlanguage[2]{{%
\expandafter\ifx\csname l@#1\endcsname\relax
\typeout{** WARNING: IEEEtran.bst: No hyphenation pattern has been}%
\typeout{** loaded for the language `#1'. Using the pattern for}%
\typeout{** the default language instead.}%
\else
\language=\csname l@#1\endcsname
\fi
#2}}

\bibitem{Zeng2018}
A.~Zeng, S.~Song, K.-T. Yu, E.~Donlon, F.~Hogan, M.~Bauza, D.~Ma, O.~Taylor,
  M.~Liu, E.~Romo, N.~Fazeli, F.~Alet, N.~Chavan-Dafle, R.~Holladay, I.~Morona,
  P.~Q. Nair, D.~Green, I.~Taylor, W.~Liu, T.~Funkhouser, and A.~Rodriguez,
  ``{Robotic Pick-and-Place of Novel Objects in Clutter with Multi-Affordance
  Grasping and Cross-Domain Image Matching},'' in \emph{IEEE International
  Conference on Robotics and Automation (ICRA)}, 2018.

\bibitem{mahler2019learning}
J.~Mahler, M.~Matl, V.~Satish, M.~Danielczuk, B.~DeRose, S.~McKinley, and
  K.~Goldberg, ``Learning ambidextrous robot grasping policies,'' \emph{Science
  Robotics}, vol.~4, no.~26, p. eaau4984, 2019.

\bibitem{visualdynamics16}
T.~Xue, J.~Wu, K.~L. Bouman, and W.~T. Freeman, ``Visual dynamics:
  Probabilistic future frame synthesis via cross convolutional networks,'' in
  \emph{Advances In Neural Information Processing Systems}, 2016.

\bibitem{ha2020fit2form}
H.~Ha, S.~Agrawal, and S.~Song, ``Fit2form: 3d generative model for robot
  gripper form design,'' \emph{arXiv preprint arXiv:2011.06498}, 2020.

\bibitem{song2020grasping}
S.~Song, A.~Zeng, J.~Lee, and T.~Funkhouser, ``Grasping in the wild: Learning
  6dof closed-loop grasping from low-cost demonstrations,'' \emph{Robotics and
  Automation Letters}, 2020.

\bibitem{redmon2015real}
J.~Redmon and A.~Angelova, ``Real-time grasp detection using convolutional
  neural networks,'' in \emph{ICRA}, 2015.

\bibitem{pinto2016supersizing}
L.~Pinto and A.~Gupta, ``Supersizing self-supervision: Learning to grasp from
  50k tries and 700 robot hours,'' in \emph{ICRA}, 2016.

\bibitem{gualtieri2016high}
M.~Gualtieri, A.~Ten~Pas, K.~Saenko, and R.~Platt, ``High precision grasp pose
  detection in dense clutter,'' in \emph{2016 IEEE/RSJ International Conference
  on Intelligent Robots and Systems (IROS)}.\hskip 1em plus 0.5em minus
  0.4em\relax IEEE, 2016, pp. 598--605.

\bibitem{lu2018planning}
Q.~Lu, K.~Chenna, B.~Sundaralingam, and T.~Hermans, ``Planning multi-fingered
  grasps as probabilistic inference in a learned deep network,'' in
  \emph{Int’l Symp. on Robotics Research}, 2017.

\bibitem{mousavian20196}
A.~Mousavian, C.~Eppner, and D.~Fox, ``6-dof graspnet: Variational grasp
  generation for object manipulation,'' in \emph{Proceedings of the IEEE
  International Conference on Computer Vision}, 2019, pp. 2901--2910.

\bibitem{gualtieri2018learning}
M.~Gualtieri and R.~Platt, ``Learning 6-dof grasping and pick-place using
  attention focus,'' in \emph{Proceedings of 2nd Conference on Robot Learning
  (CoRL 2018)}, 2018.

\bibitem{morrison2018closing}
D.~Morrison, P.~Corke, and J.~Leitner, ``Closing the loop for robotic grasping:
  A real-time, generative grasp synthesis approach,'' \emph{RSS}, 2018.

\bibitem{choi2018softrobothand}
C.~{Choi}, W.~{Schwarting}, J.~{DelPreto}, and D.~{Rus}, ``Learning object
  grasping for soft robot hands,'' \emph{IEEE Robotics and Automation Letters},
  vol.~3, no.~3, pp. 2370--2377, 2018.

\bibitem{yan2018geometryaware}
X.~{Yan}, J.~{Hsu}, M.~{Khansari}, Y.~{Bai}, A.~{Pathak}, A.~{Gupta},
  J.~{Davidson}, and H.~{Lee}, ``Learning 6-dof grasping interaction via deep
  geometry-aware 3d representations,'' in \emph{2018 IEEE International
  Conference on Robotics and Automation (ICRA)}, 2018, pp. 3766--3773.

\bibitem{liang2019pointnetgpd}
H.~Liang, X.~Ma, S.~Li, M.~G{\"o}rner, S.~Tang, B.~Fang, F.~Sun, and J.~Zhang,
  ``{PointNetGPD}: Detecting grasp configurations from point sets,'' in
  \emph{IEEE International Conference on Robotics and Automation (ICRA)}, 2019.

\bibitem{murali2020clutter}
A.~{Murali}, A.~{Mousavian}, C.~{Eppner}, C.~{Paxton}, and D.~{Fox}, ``6-dof
  grasping for target-driven object manipulation in clutter,'' in \emph{2020
  IEEE International Conference on Robotics and Automation (ICRA)}, 2020, pp.
  6232--6238.

\bibitem{zeng2016multi}
A.~Zeng, K.-T. Yu, S.~Song, D.~Suo, E.~Walker~Jr, A.~Rodriguez, and J.~Xiao,
  ``Multi-view self-supervised deep learning for 6d pose estimation in the
  amazon picking challenge,'' in \emph{Proceedings of the IEEE International
  Conference on Robotics and Automation}, 2017.

\bibitem{mahler2018dex}
J.~Mahler, M.~Matl, X.~Liu, A.~Li, D.~Gealy, and K.~Goldberg, ``Dex-net 3.0:
  Computing robust vacuum suction grasp targets in point clouds using a new
  analytic model and deep learning,'' in \emph{2018 IEEE International
  Conference on Robotics and Automation (ICRA)}.\hskip 1em plus 0.5em minus
  0.4em\relax IEEE, 2018, pp. 1--8.

\bibitem{varley2015generating}
J.~Varley, J.~Weisz, J.~Weiss, and P.~Allen, ``Generating multi-fingered
  robotic grasps via deep learning,'' in \emph{2015 IEEE/RSJ international
  conference on intelligent robots and systems (IROS)}.\hskip 1em plus 0.5em
  minus 0.4em\relax IEEE, 2015, pp. 4415--4420.

\bibitem{veres2017modeling}
M.~Veres, M.~Moussa, and G.~W. Taylor, ``Modeling grasp motor imagery through
  deep conditional generative models,'' \emph{IEEE Robotics and Automation
  Letters}, vol.~2, no.~2, pp. 757--764, 2017.

\bibitem{le2010learning}
Q.~V. Le, D.~Kamm, A.~F. Kara, and A.~Y. Ng, ``Learning to grasp objects with
  multiple contact points,'' in \emph{2010 IEEE International Conference on
  Robotics and Automation}.\hskip 1em plus 0.5em minus 0.4em\relax IEEE, 2010,
  pp. 5062--5069.

\bibitem{varley2017shape}
J.~Varley, C.~DeChant, A.~Richardson, J.~Ruales, and P.~Allen, ``Shape
  completion enabled robotic grasping,'' in \emph{2017 IEEE/RSJ international
  conference on intelligent robots and systems (IROS)}.\hskip 1em plus 0.5em
  minus 0.4em\relax IEEE, 2017, pp. 2442--2447.

\bibitem{mahler2016dex}
J.~Mahler, F.~T. Pokorny, B.~Hou, M.~Roderick, M.~Laskey, M.~Aubry,
  K.~Kohlhoff, T.~Kr{\"o}ger, J.~Kuffner, and K.~Goldberg, ``Dex-net 1.0: A
  cloud-based network of 3d objects for robust grasp planning using a
  multi-armed bandit model with correlated rewards,'' in \emph{2016 IEEE
  international conference on robotics and automation (ICRA)}.\hskip 1em plus
  0.5em minus 0.4em\relax IEEE, 2016, pp. 1957--1964.

\bibitem{shao2020unigrasp}
L.~Shao, F.~Ferreira, M.~Jorda, V.~Nambiar, J.~Luo, E.~Solowjow, J.~A. Ojea,
  O.~Khatib, and J.~Bohg, ``Unigrasp: Learning a unified model to grasp with
  multifingered robotic hands,'' \emph{IEEE Robotics and Automation Letters},
  vol.~5, no.~2, pp. 2286--2293, 2020.

\bibitem{newcombe2011kinectfusion}
R.~A. Newcombe, S.~Izadi, O.~Hilliges, D.~Molyneaux, D.~Kim, A.~J. Davison,
  P.~Kohi, J.~Shotton, S.~Hodges, and A.~Fitzgibbon, ``Kinectfusion: Real-time
  dense surface mapping and tracking,'' in \emph{2011 10th IEEE International
  Symposium on Mixed and Augmented Reality}.\hskip 1em plus 0.5em minus
  0.4em\relax IEEE, 2011, pp. 127--136.

\bibitem{Zeng2018a}
A.~Zeng, S.~Song, S.~Welker, J.~Lee, A.~Rodriguez, and T.~Funkhouser,
  ``{Learning Synergies between Pushing and Grasping with Self-supervised Deep
  Reinforcement Learning},'' in \emph{IEEE/RSJ International Conference on
  Intelligent Robots and Systems (IROS)}, 2018.

\bibitem{coumans2016pybullet}
E.~Coumans and Y.~Bai, ``Pybullet, a python module for physics simulation for
  games, robotics and machine learning,'' 2016.

\bibitem{mahler2017dex}
J.~Mahler, J.~Liang, S.~Niyaz, M.~Laskey, R.~Doan, X.~Liu, J.~A. Ojea, and
  K.~Goldberg, ``Dex-net 2.0: Deep learning to plan robust grasps with
  synthetic point clouds and analytic grasp metrics,'' \emph{RSS}, 2017.

\bibitem{toolchanger}
\BIBentryALTinterwordspacing
A.~industrial Automation, ``Automatic / robotic tool changers,'' -. [Online].
  Available: \url{http://engineering.purdue.edu/~mark/puthesis}
\BIBentrySTDinterwordspacing

\bibitem{calli2017yale}
B.~Calli, A.~Singh, J.~Bruce, A.~Walsman, K.~Konolige, S.~Srinivasa, P.~Abbeel,
  and A.~M. Dollar, ``Yale-cmu-berkeley dataset for robotic manipulation
  research,'' \emph{The International Journal of Robotics Research}, vol.~36,
  no.~3, pp. 261--268, 2017.

\end{thebibliography}

\newpage
\onecolumn

\renewcommand{\thesection}{A.\arabic{section}}
\renewcommand{\thefigure}{A\arabic{figure}}
\renewcommand{\thetable}{A\arabic{table}}
\setcounter{section}{0}
\setcounter{figure}{0}
\setcounter{table}{0}

\renewcommand{\arraystretch}{1.2}
\begin{sidewaystable}
    \centering
    \caption{Grasp Success Rate of each gripper}
   \begin{tabular}{c|c|c|cccccccc|cccccc}
    \toprule
    
    \multirow{3}{*}{Scene} &
    \multirow{3}{*}{Algorithm} &
    \multirow{3}{*}{Object} & 
    \multicolumn{8}{c|}{Training Grippers} & 
    \multicolumn{6}{c}{Testing Grippers} \\
    
    \cline{4-11} \cline{12-17}
      & & &
      \makecell[c]{WSG\\32} & Sawyer & Franka & \makecell[c]{Robotiq\\ 2F-140} & EZGripper & \makecell[c]{Kinova\\KG-3} & \makecell[c]{Robotiq\\ 3F} & \textbf{Avg.} &
      \makecell[c]{WSG\\50} & RG2 & \makecell[c]{Robotiq\\ 2F-85} & \makecell[c]{Barrett\\Hand-B} & \makecell[c]{Barrett\\Hand} & \textbf{Avg.}\\
      
    \hline
    \multirow{10}{1.1cm}{Single Object} & \multirow{2}{*}{SceneOnly} 
      & Train & 0.360 & 0.340 & 0.710 & 0.820 & 0.740 & 0.600 & 0.720 & \textbf{0.613} & 0.750 & 0.720 & 0.670 & 0.730 & 0.780 & \textbf{0.730}\\
    & & Test & 0.380 & 0.330 & 0.530 & 0.860 & 0.820 & 0.620 & 0.630 & \textbf{0.596} & 0.680 & 0.560 & 0.640 & 0.830 & 0.720 & \textbf{0.686}\\
    \cline{2-17}
    & \multirow{2}{*}{\makecell[c]{AdaGrasp\\-initOnly}} 
      & Train & 0.600 & 0.530 & 0.800 & 0.730 & 0.850 & 0.710 & 0.830 & \textbf{0.721} & 0.900 & 0.770 & 0.640 & 0.770 & 0.880 & \textbf{0.792}\\
    & & Test & 0.610 & 0.490 & 0.690 & 0.740 & 0.860 & 0.790 & 0.850 & \textbf{0.719} & 0.860 & 0.650 & 0.660 & 0.810 & 0.785 & \textbf{0.753}\\
    \cline{2-17}
    & \multirow{2}{*}{\makecell[c]{AdaGrasp\\-fixConfig}} 
      & Train & 0.650 & 0.570 & 0.810 & 0.890 & 0.810 & 0.800 & 0.830 & \textbf{0.766} & 0.910 & 0.840 & 0.810 & 0.870 & 0.845 & \textbf{0.855}\\
    & & Test & 0.610 & 0.500 & 0.780 & 0.950 & 0.890 & 0.810 & 0.850 & \textbf{0.770} & 0.880 & 0.890 & 0.780 & 0.870 & 0.850 & \textbf{0.854}\\
    \cline{2-17}
    & \multirow{2}{*}{\makecell[c]{AdaGrasp\\-fixGripper}} 
      & Train & 0.890 & 0.990 & 0.980 & 0.960 & 0.900 & 0.920 & 0.820 & \textbf{0.923} & 1.000 & 0.890 & 0.910 & 0.910 & 0.815 & \textbf{0.905}\\
    & & Test & 0.950 & 0.980 & 1.000 & 1.000 & 0.920 & 0.970 & 0.890 & \textbf{0.959} & 0.990 & 0.940 & 0.970 & 0.940 & 0.850 & \textbf{0.938}\\
    \cline{2-17}
    & \multirow{2}{*}{AdaGrasp} 
      & Train & - & - & - & - & - & - & - & \textbf{0.960} & - & - & - & - & - & \textbf{1.000}\\
    & & Test & - & - & - & - & - & - & - & \textbf{0.970} & - & - & - & - & - & \textbf{0.990}\\

    \hline\hline
    \multirow{10}{1.1cm}{Multi Object} & \multirow{2}{*}{SceneOnly} 
      & Train & 0.324 & 0.364 & 0.596 & 0.688 & 0.668 & 0.428 & 0.380 & \textbf{0.493} & 0.676 & 0.584 & 0.580 & 0.348 & 0.450 & \textbf{0.528}\\
    & & Test & 0.328 & 0.320 & 0.484 & 0.736 & 0.728 & 0.496 & 0.384 & \textbf{0.497} & 0.580 & 0.572 & 0.576 & 0.456 & 0.472 & \textbf{0.531}\\
    \cline{2-17}
    & \multirow{2}{*}{\makecell[c]{AdaGrasp\\-initOnly}} 
      & Train & 0.556 & 0.504 & 0.664 & 0.656 & 0.752 & 0.616 & 0.712 & \textbf{0.637} & 0.760 & 0.652 & 0.532 & 0.648 & 0.776 & \textbf{0.674}\\
    & & Test & 0.544 & 0.440 & 0.568 & 0.732 & 0.788 & 0.672 & 0.724 & \textbf{0.638} & 0.700 & 0.484 & 0.504 & 0.712 & 0.730 & \textbf{0.626}\\
    \cline{2-17}
    & \multirow{2}{*}{\makecell[c]{AdaGrasp\\-fixConfig}} 
      & Train & 0.596 & 0.504 & 0.724 & 0.904 & 0.784 & 0.664 & 0.764 & \textbf{0.706} & 0.780 & 0.824 & 0.684 & 0.688 & 0.778 & \textbf{0.751}\\
    & & Test & 0.528 & 0.440 & 0.692 & 0.904 & 0.864 & 0.684 & 0.684 & \textbf{0.685} & 0.804 & 0.800 & 0.684 & 0.760 & 0.770 & \textbf{0.764}\\
    \cline{2-17}
    & \multirow{2}{*}{\makecell[c]{AdaGrasp\\-fixGripper}} 
      & Train & 0.860 & 0.912 & 0.936 & 0.852 & 0.852 & 0.820 & 0.712 & \textbf{0.849} & 0.916 & 0.892 & 0.856 & 0.760 & 0.784 & \textbf{0.842}\\
    & & Test & 0.884 & 0.916 & 0.956 & 0.912 & 0.848 & 0.904 & 0.704 & \textbf{0.875} & 0.940 & 0.876 & 0.880 & 0.796 & 0.780 & \textbf{0.854}\\
    \cline{2-17}
    & \multirow{2}{*}{AdaGrasp} 
      & Train & - & - & - & - & - & - & - & \textbf{0.912} & - & - & - & - & - & \textbf{0.908}\\
    & & Test & - & - & - & - & - & - & - & \textbf{0.896} & - & - & - & - & - & \textbf{0.936}\\

    \hline \hline
    \multirow{10}{1.1cm}{Multi Object with Obstacle} & \multirow{2}{*}{SceneOnly} 
      & Train & 0.193 & 0.193 & 0.400 & 0.513 & 0.507 & 0.307 & 0.313 & \textbf{0.347} & 0.367 & 0.420 & 0.400 & 0.293 & 0.360 & \textbf{0.368}\\
    & & Test & 0.133 & 0.140 & 0.260 & 0.433 & 0.407 & 0.267 & 0.260 & \textbf{0.271} & 0.353 & 0.300 & 0.267 & 0.307 & 0.287 & \textbf{0.303}\\
    \cline{2-17}
    & \multirow{2}{*}{\makecell[c]{AdaGrasp\\-initOnly}} 
      & Train & 0.453 & 0.347 & 0.613 & 0.673 & 0.533 & 0.440 & 0.400 & \textbf{0.494} & 0.253 & 0.433 & 0.460 & 0.360 & 0.257 & \textbf{0.353}\\
    & & Test & 0.447 & 0.333 & 0.580 & 0.753 & 0.647 & 0.367 & 0.467 & \textbf{0.513} & 0.260 & 0.360 & 0.487 & 0.353 & 0.253 & \textbf{0.343}\\
    \cline{2-17}
    & \multirow{2}{*}{\makecell[c]{AdaGrasp\\-fixConfig}} 
      & Train & 0.487 & 0.480 & 0.700 & 0.753 & 0.687 & 0.587 & 0.593 & \textbf{0.612} & 0.620 & 0.673 & 0.580 & 0.247 & 0.493 & \textbf{0.523}\\
    & & Test & 0.480 & 0.387 & 0.633 & 0.773 & 0.733 & 0.587 & 0.627 & \textbf{0.603} & 0.640 & 0.720 & 0.607 & 0.340 & 0.537 & \textbf{0.569}\\
    \cline{2-17}
    & \multirow{2}{*}{\makecell[c]{AdaGrasp\\-fixGripper}} 
      & Train & 0.813 & 0.860 & 0.873 & 0.693 & 0.720 & 0.780 & 0.687 & \textbf{0.775} & 0.780 & 0.760 & 0.693 & 0.427 & 0.630 & \textbf{0.658}\\
    & & Test & 0.847 & 0.867 & 0.867 & 0.793 & 0.780 & 0.820 & 0.720 & \textbf{0.813} & 0.787 & 0.800 & 0.787 & 0.473 & 0.667 & \textbf{0.703}\\
    \cline{2-17}
    & \multirow{2}{*}{AdaGrasp} 
      & Train & - & - & - & - & - & - & - & \textbf{0.853} & - & - & - & - & - & \textbf{0.747}\\
    & & Test & - & - & - & - & - & - & - & \textbf{0.887} & - & - & - & - & - & \textbf{0.793}\\
    \bottomrule
    \end{tabular}
\end{sidewaystable}

\end{document}